# Mnesors for Automatic Control

Gilles CHAMPENOIS

*Collège Saint-André, Saint-Maur, France*
*gilles_champenois@yahoo.fr*

ABSTRACT. Mnesors are defined as elements of a semimodule over the min-plus integers. This two-sorted structure is able to merge graduation properties of vectors and idempotent properties of boolean numbers, which makes it appropriate for hybrid systems. We apply it to the control of an inverted pendulum and design a full logical controller, that is, without the usual algebra of real numbers.

> *"Thus the theory of control in engineering, whether human or animal or mechanical, is a chapter in the theory of messages."*
>
> *Norbert Wiener (1894-1964)*

## I. INTRODUCTION

Machines generally must be controlled or constrained to achieve a certain goal by a human pilot or by an automat which is called a controller. When constant action of the pilot is required, we call it continous control or, in the case of the automat, feedback control. For example, the cars can have their speed kept more or less constant by the driver or by an electronic controller.

But to those basic tasks, feedback controllers have added new functionalities. They generally take over supervision tasks. For instance, some car electronic systems can alert on a possible danger of collision. That is what one calls discrete event control.

Unfortunately, the two cases, continous and discrete control, are based on two different kinds of computing. The former is based on real number computing, the latter on boolean numbers, and each with completely different mathematical operations. Still they have to interact. So it is clearly needed to find a common standard, that is, common computing operations on common numbers.

Although software could help by putting both continous and discrete variables in boolean format (coded by 0 or 1), their mixture is clearly artificial. In fact it brings nothing to help the design. Basically, variables are of completely different meaning. So we must say that numerical technology fails to fill the gap here.

Control theory alone is made to do the job. For that, we must find some control laws that can combine continous and discrete variables into a single format with common operations. Fuzzy control [1] positively made the way towards a unified control by dropping the regular algebra of real numbers and instead by using logical rules of numerical calculus. We want to continue the work and propose a logical controler based on the mnesor calculus (mnesor controller). Its main advantage is to merge linear properties of vectors and boolean properties.

Again we see a number of reasons to apply this calculus to automatic controllers: full compatibility between discrete and continous control, minimized computing resources, linear control. All this makes it appropriate tools for hybrid systems and HMI systems.

We'll finish with the inverted pendulum example.

## II. THE MIN-PLUS INTEGERS

The usual operations on integers are the addition ($+$) and multiplication ($\times$). Still we want to reconsider that with a different pair of operations (written $\oplus, \otimes$): the minimum and the addition:

$$x \oplus y = \min(x, y) \text{ and}$$
$$x \otimes y = x + y.$$

But for the sake of simplicity they are respectively renamed addition and multiplication.

EXAMPLE. $1 \oplus 2 = 1$
$2 \otimes -3 = -1$

Such numbers are called flat numbers and are written $(0), (+1), (+2)$, etc. and for the negative integers: $(0), (-1), (-2)$ etc. $(-1)$ is also written $(+1)^{-1}$. Their set is denoted by $G°$. Whenever the sum $x \oplus y$ is equal to $y$ itself, we say that $x$ is inferior to $y$. Thus compared to the standard ordering the order here is reverse. Unlike standard ordering where $(-1) \leq (0) \leq (+1)$, positive numbers on the scale are given the lower position and we can write $(-1) \succ (0) \succ (+1)$ as it is shown graphically below:

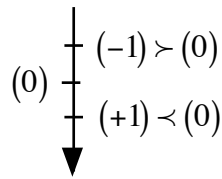

FIGURE 1. The ordering of flat numbers

## III. FLAT MNESORS

Let $M°$ be a commutative monoid with an identity element $0 \in M°$. We assume that there exists an external multiplication (for all $x \in M°$ and $\lambda \in G°$, $x\lambda \in M°$) with four properties (see properties 1-2-3-4 below).

$$x(0) = x \tag{1}$$
$$(x + y)\lambda = x\lambda + y\lambda \tag{2}$$
$$(x\lambda)\mu = x(\lambda \otimes \mu) \tag{3}$$
$$x(\lambda \oplus \mu) = x\lambda + x\mu \tag{4}$$

That's formally what defines a semimodule over the flat numbers. The elements are named mnesors and precisely flat mnesors.

Furthermore, we add a unary function ($x \to \bar{x}$ the conjugate of $x$) which is:

- involutive, $\bar{\bar{x}} = x$ (5)
- compatible with the external multiplication, $\overline{x\lambda} = \bar{x}\lambda^{-1}$ (6)

*Idempotence.* We prove first that the addition of mnesors is idempotent

PROOF. In $x + x$, we substitute $x\tau$ for $x$. Thus $x + x = x\tau + x\tau = x(\tau \oplus \tau) = x\tau = x$

*Ordering.* The ordering is defined as follows: the mnesor $x$ is inferior to the mnesor $y$ (written $x \leq y$) if and only if $x$ can be written as $x = y\lambda$ where $\lambda \prec (0)$.

PROOF. Obviously reflexive and transitive.
Antisymmetric: $x \leq y$ and $y \leq x$, that is $x = y\lambda$ and $y = x\mu$. Then
$x + y = x + x\mu = x(0) + x\mu = x((0) \oplus \mu) = x(0) = x$ and $x + y = y\lambda + y = y$. Then $x = y$.

We prove next that for all $x$ in $G°$, $x\lambda \leq x\mu$ whenever $\lambda \prec \mu$.

PROOF. As the addition in $G°$ returns one of the operands, for example we can write $\lambda \oplus \mu = \mu$.
So, $x\lambda + x\mu = x(\lambda \oplus \mu) = x\mu$

*Empty mnesor.* The mnesor identity $0$ is called the empty mnesor and it verifies the identity $0\lambda = 0$ whenever $\lambda \prec (0)$

PROOF. $0\lambda = 0 + 0\lambda = 0(0) + 0\lambda = 0((0) \oplus \lambda) = 0(0) = 0$

*(Internal) multiplication.* The product of the mnesors $x$ and $y$ is the mnesor $x \times y$ defined as $x \times y = \bar{x} + \bar{y}$. The operation is associative, commutative and idempotent. Two properties hold:

- $(x \times y)\lambda = (x\lambda) \times (y\lambda)$,
- $x \times (x\lambda) = x\lambda$, with $\lambda \prec (0)$

PROOF. $(x \times y)\lambda = \overline{(\bar{x} + \bar{y})}\lambda = \overline{(\bar{x} + \bar{y})\lambda^{-1}} = \overline{\bar{x}\lambda^{-1} + \bar{y}\lambda^{-1}} = \overline{\overline{x\lambda} + \overline{y\lambda}} = (x\lambda) \times (y\lambda)$

$$x \times (x\,\lambda) = \overline{\overline{x} + \overline{x}\lambda^{-1}} = \overline{\overline{x}\bigl((0) + \lambda^{-1}\bigr)} = x\,\lambda$$

## IV. NUMBERS FOR QUALITATIVE REASONING

As far as the usual algebra holds, a number is without ambiguity either positive or negative. But if we now consider AI, its status is not that clear. It can be more or less positive, larger or not, more or less precise. Isn't 2,00 € more precise than two euros? Following an example from [1], isn't the number 1 less positive than the number 10? More generally, numbers appear with fuzzy qualities: being positive, negative, large, centered, precise, etc. We assume here that there exist two fuzzy sets, *PST* and *NGT*, one for positivity and the other one for negativity.

Let's now give an interpretation of the external multiplication. We have seen that *PST* expresses positivity. *PST* multiplied by 2 expresses positivity too but with a stronger constraint. Generally, the larger the integer, the less *PST* multiplied by it will be matched by a given number, or equivalently, the less positive this number will appear. For example, if 10 matches *PST* well, it matches *PST* times 2 less because the positivity constraint is higher in *PST* times 2 than in *PST* alone.

Now what do the addition and the (internal) multiplication of mnesors mean? Adding mnesors relaxes each operand by simply returning the mnesor with the softer constraint. For example, the addition of *PST* to *PST* times 2 returns *PST* which is the less constrained mnesor. Compared with the addition, the multiplication has the opposite effect by returning a more constraint mnesor. For instance, the product of *PST* by *PST* times 2 is the more constrained mnesor of both, that is the latter.

Within the general framework of mnesors, we add a couple of rules:

Every flat mnesor is linear with *PST*, *NGT* or $ALL = \overline{0}$

$$\text{☑}\ x = PST\,\lambda \ \text{or}\ x = NGT\,\lambda \ \text{or}\ x = ALL\,\lambda \tag{7}$$

The conjugate of *PST* is *NGT* and vice-versa,

$$\text{☑}\ \overline{PST} = NGT \tag{8}$$

The sum of a mnesor inferior or equal to $x$ to the supplement of $x$,

$$\text{☑}\ x\,\lambda + \overline{x} = \overline{x} \qquad \text{for } \lambda \prec (0) \text{ but } \lambda \neq (0) \tag{9}$$
$$\text{☑}\ x + \overline{x} = ALL \tag{10}$$

Further rules can be deduced:

- $x + ALL = ALL$
- $x \times ALL = x$,
- $x \times 0 = 0$,
- $x \times \overline{x} = 0$,
- $(x\lambda) \times (x\mu) = x\min(\lambda,\mu)$.

PROOFS.

$x + ALL = x + x + \overline{x} = x + \overline{x} = ALL$

$x \times ALL = \overline{\overline{x} + \overline{ALL}} = \overline{\overline{x} + 0} = \overline{\overline{x}} = x$

$x \times 0 = \overline{\overline{x} + \overline{0}} = \overline{\overline{x} + ALL} = \overline{ALL} = 0$

$x \times \overline{x} = \overline{\overline{x} + \overline{\overline{x}}} = \overline{\overline{x} + x} = \overline{ALL} = 0$

$x\lambda \times x\mu = \overline{\overline{x\lambda} + \overline{x\mu}} = \overline{\overline{x}(\lambda^{-1} \oplus \mu^{-1})} = \overline{x}\left(\lambda^{-1} \oplus \mu^{-1}\right)^{-1} = x\min(\lambda,\mu)$

## V. AN APPLICATION TO AUTOMATIC CONTROL

Our inverted pendulum is made of a mass point (mass $m$) at the upper extremity of a vertical weightless bar (length $a$). A cart (position $x$) moves the other extremity on the x-axis and help it to return back to the vertical position against gravity ($g$). $\theta$ and $\omega$ are respectively the angle of the bar with the vertical and its rate (see below).

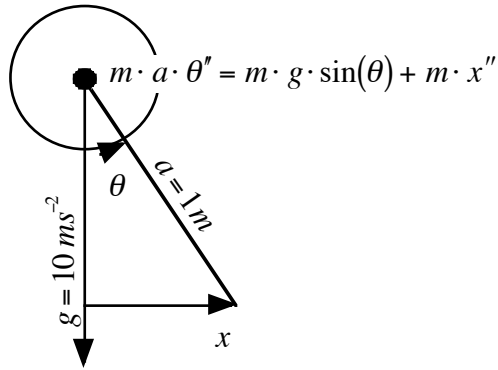

FIGURE 2. A simple inverted pendulum

We have to stabilize the pendulum around the vertical, that is, to keep both variables $\theta$ and $\omega$ close to zero through the direct control of $x''$ (as control variable, $x''$ is renamed $u$). We want to compare two different control laws: the first one (P(I)D law) is based on the usual algebra and is simply computed as $u = -10\theta - 5\omega$ (bound to the $\pm 30\,ms^{-2}$ interval). The other one is based on the mnesor theory.

We first performed the simulation of the pendulum controled by the P(I)D law. The error, rate and control signals are then pictured below:

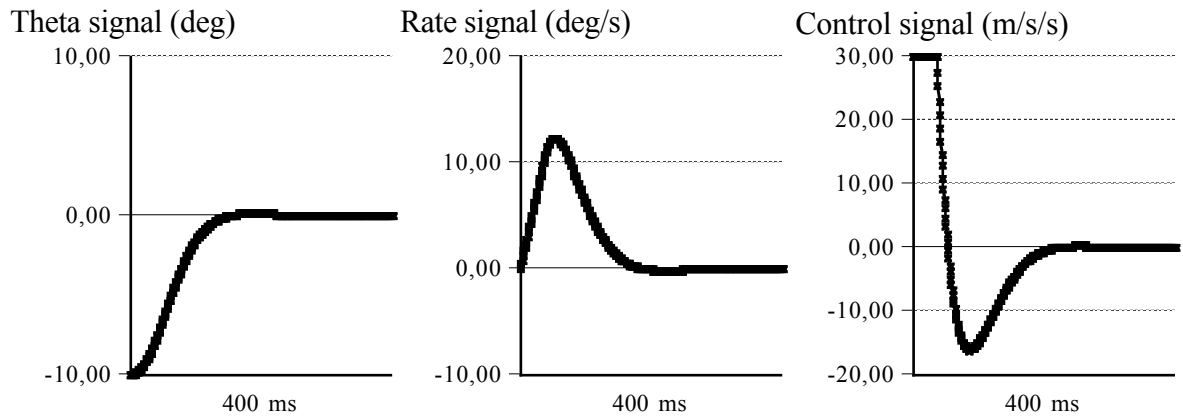

FIGURE 3. A P(I)D control law

Now we want to compare with the mnesor control law. First converting values of $u,\theta,\omega$ into three mnesors ($U,\Xi,\Omega$) is needed. The mnesor $\Xi$ (resp. $\Omega$) is equal to $PST\,\lambda$ whenever $\theta$ is positive and is equal to $NGT\,\lambda$ whenever $\theta$ is negative. The more $\theta$ (resp. $\omega$) is, the less $\lambda$ is. In details have a look at the diagram below.

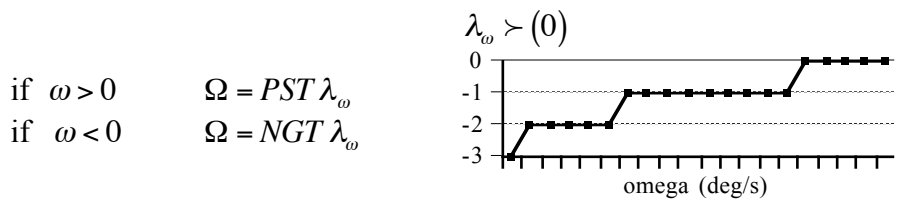

if $\omega > 0$    $\Omega = PST\,\lambda_\omega$
if $\omega < 0$    $\Omega = NGT\,\lambda_\omega$

FIGURE 4. $\omega$-to-$\Omega$ converter

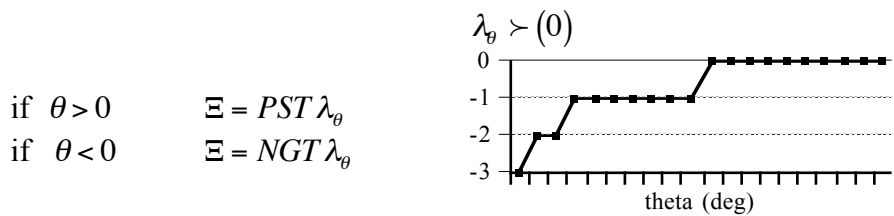

if $\theta > 0$    $\Xi = PST\,\lambda_\theta$
if $\theta < 0$    $\Xi = NGT\,\lambda_\theta$

FIGURE 5. $\theta$-to-$\Xi$ converter

If $u,\theta,\omega$ are positive (resp. negative), then $U,\Xi,\Omega$ are given $PST$ ($NGT$) mnesor. We recall that multiplying two mnesors means choosing the more constrained of both ($PST \times NGT\,\lambda = PST$ and

$NGT \times PST\, \lambda = NGT$ if $\lambda \succ (0)$). If one is as positive but the other one is negative, the multiplication returns zero ($PST \times NGT = 0$). So whenever theta and rate are of different signs, the control is zero.

The mnesor control $U$ is calculated as follows: $U = \Xi \times \Omega$

Then it is converted back to real values (see below).

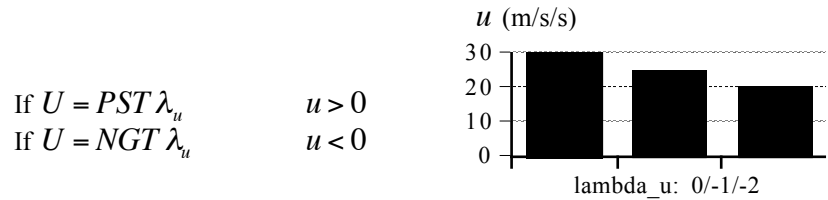

If $U = PST\, \lambda_u$     $u > 0$
If $U = NGT\, \lambda_u$     $u < 0$

FIGURE 6. $U$-to-$u$ converter

The error, rate and control signals are pictured below:

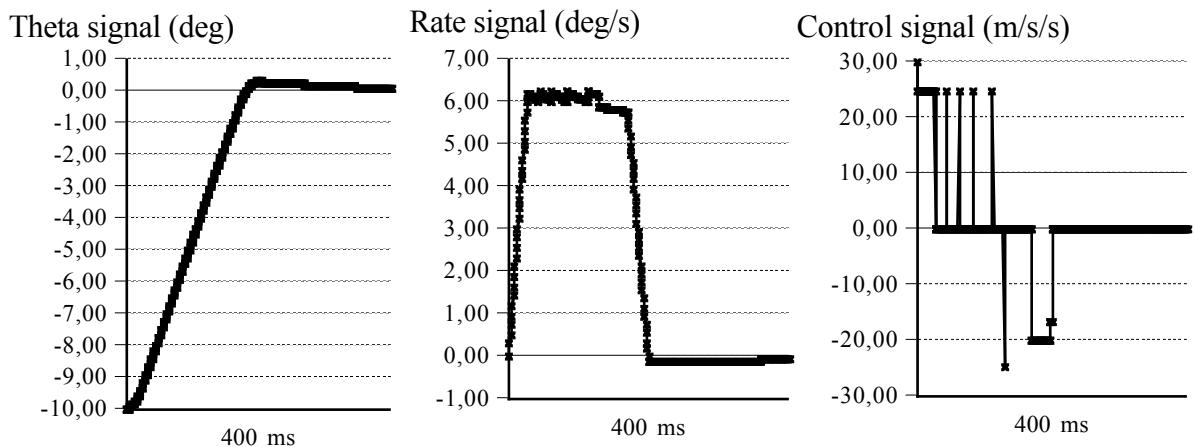

FIGURE 7. Mnesor control law

## CONCLUSION

One can notice the similarity to fuzzy control. But here the theory is completely axiomatized. Fuzzification (what we call real-to-mnesor conversion) seems more natural and defuzzification much more simple.

## REFERENCES

1. ZADEH L. (1965), *Fuzzy Sets*